# PALRACE: Reading Comprehension Dataset with Human Data and Labeled Rationales

Jiajie Zou[2], Yuran Zhang[2], Peiqing Jin[2], Cheng Luo[2], Xunyi Pan[3], Nai Ding[1,2*]

[1]Zhejiang Lab / Hangzhou, China

[2]Key Laboratory for Biomedical Engineering of Ministry of Education, College of Biomedical Engineering and Instrument Sciences, Zhejiang University / Hangzhou, China

[3]School of International Studies, Zhejiang University / Hangzhou, China

## Abstract

Pre-trained language models achieves high performance on machine reading comprehension (MRC) tasks but the results are hard to explain. An appealing approach to make models explainable is to provide rationales for its decision. To investigate whether human rationales can further improve current models and to facilitate supervised learning of human rationales, here we present PALRACE (Pruned And Labeled RACE), a new MRC dataset with human labeled rationales for 800 passages selected from the RACE dataset. We further classified the question to each passage into 6 types. Each passage was read by at least 26 human readers, who labeled their rationales to answer the question. It is demonstrated that models such as RoBERTa-large outperforms human readers in all 6 types of questions, including inference questions, but its performance can be further improved when having access to the human rationales. Simpler models and pre-trained models that are not fine-tuned based on the task benefit more from human rationales, and their performance can be boosted by more than 30% by rationales. With access to human rationales, a simple model based on the GloVe word embedding can reach the performance of BERT-base.

## 1 Introduction

Recently, pre-trained language models have reached human-level performance on many MRC tasks. The high-performance of pre-trained models, however, is achieved at the cost of explainablity and robustness[1-4]. One way to improve the explainablity and robustness of models is to let the model provides rationales for its decision[5,6]. Providing rationales is especially important and challenging for MRC tasks, since it is often challenging to locate question-related information in the long passage. Recently, a number of MRC datasets with human labeled rationales have been proposed[7,8], but these datasets generally consist of a relatively small set of questions and the rationales, i.e., key sentences or continuous spans, are labeled by a few annotators.

---

[*] Corresponding author

Here we propose a new dataset that has 800 passage/question pairs. Rationales are labeled at word-level granularity by multiple annotators. Importantly, using behavioral tests, it is confirmed that the labeled rationales are of high quality and can sufficiently support question answering. In the following, it is also shown that human rationales can improve the performance of pre-trained models.

The purpose to design this dataset is twofold. First, such a dataset can support supervised learning of rationales. If a MRC model can explicitly provide rationales, on top of giving an answer, the results will be much more interpretable. Second, human rationales can be viewed as gold standard for where question-related information is located in a passage. When models have access to the human rationales, their question answering accuracy solely relies on their ability to interpret the rationales. Therefore, using human rationales, we can test whether the performance of MRC models is limited by the ability to locate information or by the ability to interpret information.

## 2  Dataset

### 2.1  Dataset collection

The reading comprehension passages and questions were selected from the test set of the large-scale RACE dataset[9]. Only the high-school level questions were selected. All reading comprehension questions were multiple-choice questions with 4 answer choices and only one choice was correct. The comprehension questions included six types, i.e., Cause, Fact, Inference, Theme, Title, and Purpose.

The first 3 types of questions were targeted to test readers' ability to locate, extract and understand specific information from a passage, and were referred to as local questions. Specifically, a Cause question asked about the cause or intention of a certain action or activity described in a passage. For example, "What is the first experiment aimed at?" A Fact question was concerned with specific facts or details stated directly in a passage. For example, "Who experienced the German occupation?" An Inference question was concerned with judgments or conclusions readers made based on the information provided in a passage. For example, "What did the writer's friend most probably complain about?"

Questions of the other three types, namely Theme, Title, and Purpose, tested the understanding of a passage in its totality, and were referred to as global questions. A uniform question stem was used for all global questions within the same type. For Theme, the question stem was: "What is this passage mainly about?" For Title, the question stem was: "What is the best title for this passage?" For Purpose, the question stem was: "What is the main purpose of the passage?"

Two graduate students majored in English literature manually selected the questions for each question types. Besides, previous studies have denoted that some questions in RACE have wrong answers[1], and some can be guessed without reading passages[1,3]. The 2 graduate students also proofread all passages, corrected occasional typos, and verified the answer to each comprehension question was correct and could not be easily guessed without reading the passage. Finally, a total of 800 questions passed the verification and were included in our dataset. The number of questions included in each question type is reported in Table 1.

## 2.2 Rationale annotation

**Participants.** Totally, 179 participants (20-34 years old, mean age, 23.8 years; 124 female) took part in the annotation experiment. The number of participants involved in the annotation of each question type can be found in Table S2, and each participant was allowed to annotate passages of several question types. All participants were Chinese undergraduate or graduate students. English proficiency of the participants were guaranteed by the following criterion: a minimum score of 6 on IELTS, 80 on TOEFL, or 425 on CET6. Participants got basic salary per question plus bonus if the question was answered correctly in the evaluation session of the experiment.

**Procedures of the annotation experiment.** In the annotation session, participants read a question and its corresponding passage, without options provided. They were required to annotate words in the passage that were relevant to answering the question. To ensure that only the most important words got annotated, each participant could only annotate up to 5 words in each passage.

In the evaluation session, conducted at least 1 week after the annotation session, participants were presented with a question, 4 candidate options and a masked passage, in which they could only see the annotated words. For words that were not annotated, each letter was replaced with a letter z. Participants were required to choose the best answer for the question based on annotated words.

Before the annotation session, participants were informed that the evaluation session was scheduled a week later, and their bonus would be calculated in proportion to their performance in question answering in the evaluation session. The evaluation session was employed to encourage the participants to annotate words that were of crucial importance to question answering, and to evaluate whether the annotated words were sufficient for the question to be answered.

Before the experiment, participants were randomly paired up. In the annotation session, all questions were randomly divided into two equal parts and each participant in a pair annotated one half of the questions. In the evaluation session, however, each participant had to answer questions annotated both by him/herself and by the partner (illustrated in Figure 1). The accuracy of answers generated in the evaluation experiment was 73.41% on average (see Table 1 for the performance of each question type), indicating that the annotation was of high quality and could sufficiently support question answering.

Since each passage was labeled by dozens of annotators (Table 1), we counted how many times a word was labeled. An example for the annotation results is shown in Figure 2. In all analysis, we defined the 5 words that were most frequently labeled as the rationales.

## 2.3 Human performance

To measure human performance, each question was answered by a new group of 25 participants based on the whole passage. Furthermore, in the rationale annotation experiment, we also tested whether participants can answer the questions by reading just the rationales (up to 5 words), instead of the passage. Results of both experiments were reported in Table 1. It was found that human performance did not strongly differ whether they read the whole passage or just the rationales, confirming that the rationales were of high quality and can sufficiently support question answering.

# 3 Experiments

## 3.1 Models

We tested our dataset using transformer-based models, including BERT, ALBERT, and RoBERTa. We tested both the base and large versions of each model. Following previous work[10,11], we constructed the input in the form of [CLS] < passage> [SEP] < question, option> [SEP] for each option. We utilized a linear feedforward layer to convert the final representations of the [CLS] token into a scalar, and then applied a softmax layer over the 4 scalars of 4 options to compute the scores of options being correct. All the models were trained to maximize the log-score of the correct option, and were implemented based on HuggingFace[12].

We fine-tuned the transformer models based on the large-scale RACE dataset. All hyperparameters for fine-tuning were adopted from previous studies (shown in Appendix A.2). For comparison, we also tested the pre-trained models in which none of the modules but the final linear feedforward layer was fine-tuned.

We also tested a more basic model based on the 100-dimensional GloVe word embeddings[13]. For this GloVe-based model, we simply summed the word embeddings across all words in a passage to get a passage embedding. For each option, we also summed the word embeddings across all words in the option to get an embedding for the option. The option whose embedding has the maximum cosine similarity with passage embedding is chosen as the answer.

### 3.2 Models with access to human rationales

Human rationales from different annotators were pooled, and 5 words most frequently labeled by the annotators were retained for further analysis. Furthermore, since context may be crucial for the transformer-based models, we constructed two types of rationales: a word-level rationale consisting of 5 frequently labeled words; a sentence-level rationale consisting of sentences in which the 5 frequently labeled words appear. For models with access to human rationales, we replaced the passage with either the word-level rationale or the sentence-level rationale.

When models answered questions based on just the rationales, we did another fine-tuning using 10-fold cross validation. In each fold, the dataset was split into 70:20:10 to generate training set, development set, and test set, respectively. Each passage appeared once in the test set across the 10 folds. The hyperparameters are reported in Appendix A. 1.

### 3.3 Performance of different models

We first analyze the performance of BERT-large, ALBERT-large, and RoBERTa-large (Table 2), which are state-of-the-art models with high performance on the RACE dataset. It is shown that these models generally have better performance for global questions than local questions, especially for BERT-large whose performance is about 10% higher for global questions. Models generally outperform humans in answering global questions but underperform humans in local questions. RoBERTa-large has the highest overall accuracy and outperforms human in all 6 question types. When having access to sentence-level human rationales, the overall performance of all 3 large transformer models is further boosted. Of the 3 models,

BERT-large shows the largest improvement and its performance increases by more than 10%. In terms of question types, local questions benefit more from human rationales.

On top of the 3 large transformer models, in the following, we further analyze the performance of fine-tuned base-version transformer models, base-version transformer models that were not fine-tuned, and a basic model based on GloVe word embeddings. Here we show the results for 3 types of questions in Figure 3, and the results for the other 3 types can be found in Figure A1. For models that have no access to human rationales (black bars in Figure 3), it is clear that no model perform well without fine-tuning. When models have access to the sentence or word rationales (blue and red bars in Figure 3), performance generally improves, especially for the local questions such as Fact and Inference questions.

### 3.4 Comparisons between models

For models that are not fine-tuned, best performance is generally observed when word-level rationales are provided. In other words, these models achieve the best performance if the input has only 5 key words. When more words are available, performance decreases, suggesting that these models have very limited ability to extract key information embedded in high dimensional data. From another perspective, the results show that fine-tuning mainly improves a model's ability to locate key information, instead of its ability to integrate a few key words to answer a multiple-choice question.

## 4 Conclusion

In sum, we propose PALRACE, a high-quality MRC dataset that includes 800 passages with human-labeled word-level rationales. Each passage was annotated by at least 26 annotators. Our experiments demonstrated that human rationales can boost the performance of MRC models, especially for models that do not achieve state-of-the-art performance without human rationales. Importantly, with rationales, fine-tuning has less influence on the model performance. In other words, models with rationales are more robust, and are less sensitive to how the models are fine-tuned. Furthermore, it is demonstrated that PALRACE could help to design interpretable models that do not just answer questions but also provide rationales.

# Tables

Table 1: Statistics of PALRACE dataset.

|  | local questions | | | global questions | | |
|---|---|---|---|---|---|---|
| Type | Cause | Fact | Inference | Theme | Title | Purpose |
| # passages | 200 | 200 | 120 | 100 | 100 | 80 |
| passage length | 141-426 | 121-397 | 117-452 | 173-397 | 163-254 | 206-456 |
| # annotators | 37 | 37 | 34 | 37 | 38 | 26 |
| accuracy (passage) | 80.00% | 78.74% | 71.23% | 79.76% | 77.80% | 78.75% |
| accuracy (rationale) | 76.28% | 79.02% | 72.38% | 70.53% | 67.25% | 65.02% |

Table 2: Performance of humans and pre-trained models for each type of questions. The questions are answered based on either the whole passage or sentence-level rationales, i.e., rationale$^S$. Numbers in bold shows whether the performance is higher for passage or rationale$^S$.

| | evidence | average | Cause | Fact | Inference | Theme | Title | Purpose |
|---|---|---|---|---|---|---|---|---|
| BERT-L | passage | 70.88% | 67.50% | 68.50% | 62.50% | **83.00%** | 78.00% | 73.75% |
| | rationale$^S$ | **81.13%** | **85.00%** | **78.50%** | **80.00%** | 80.00% | **80.00%** | **82.50%** |
| ALBERT-L | passage | 80.75% | 79.00% | 78.50% | **78.33%** | **88.00%** | **83.00%** | 82.50% |
| | rationale$^S$ | **82.62%** | **86.50%** | **81.50%** | **78.33%** | 83.00% | 80.00% | **85.00%** |
| RoBERTa-L | passage | 88.75% | 93.00% | 85.00% | 84.17% | **94.00%** | 87.00% | **90.00%** |
| | rationale$^S$ | **89.25%** | **95.00%** | **85.50%** | **87.50%** | 84.00% | **92.00%** | **90.00%** |

# Figures

Figure. 1: Procedure of the annotation experiment.

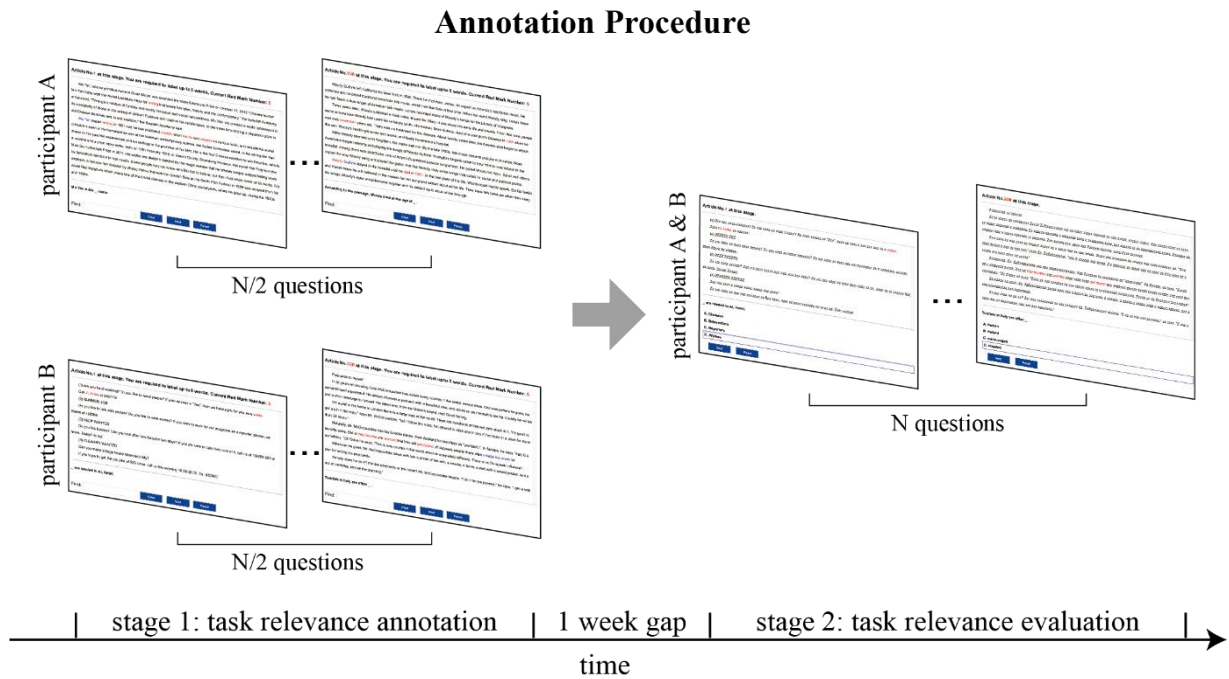

Figure. 2 Illustration of the labeled rationales. Colorbar shows the percent of annotators labeling each word.

*Passage:*
"…When the seeds sprouted, Julie looked anxiously at the rows and there were no empty spots. They all came up and produced the best crop of vegetables Bernie had ever seen. They soon became well-known in the community for having the best prices and finest produce in town. They also extended credit to their neighbors and accepted various items as trade for merchandise. The two of them worked hard throughout the 1930s and made a very successful business. Julie earned a salary. As the economy picked up, so did sales at the store..."

*Question:* We can infer from the passage that sales at the store has been _ throughout the 1930s.
*Options:*  A. fluctuating    B. increasing    C. changing    D. stable

Figure. 3 performance of different models for 3 types of questions. Models with ``-O'', ``-B'', and ``-L'' suffix denote the original (i.e., without fine-tuning), base-version, and large-version, respectively. The red dashed line and the black dashed line denotes the human performance with the full passage and 5 rationales, respectively.

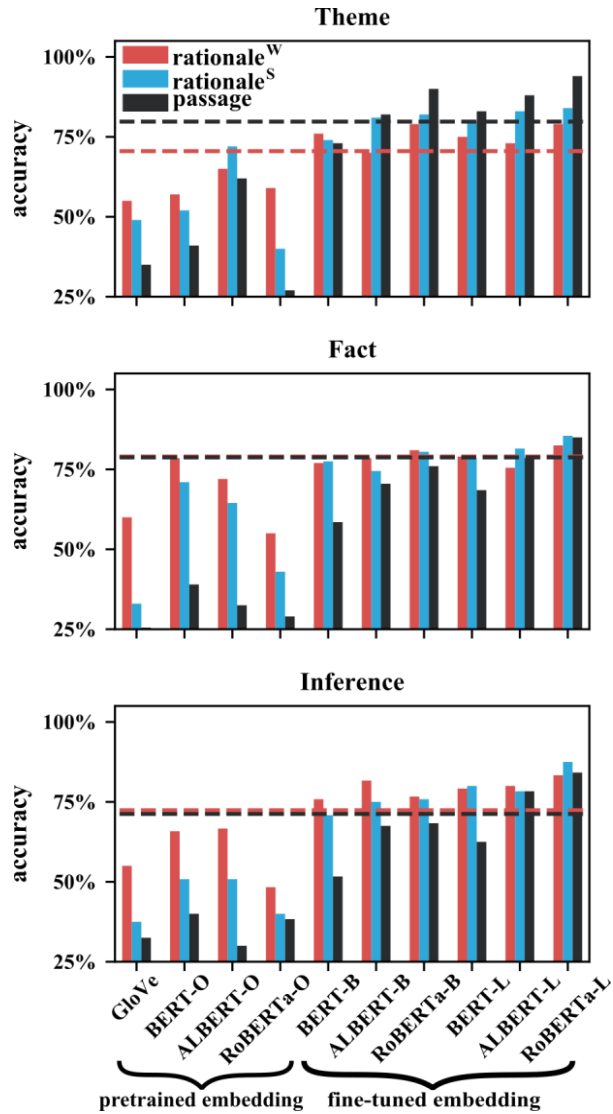

# A. Appendices

## A.1 Hyperparameters

The Hyperparameters for fine-tuning on RACE and fine-tuning for integrating with rationales are shown in Table A1 and Table A2.

Table A1: Hyperparameters for fine-tuning on RACE. We adapted these hyperparamemers from previous studies.

|  | models | learning rate | training epochs | training steps | training batch size | warmup steps | weigh decay |
|---|---|---|---|---|---|---|---|
| BERT | BERT-base | 1e-5 | 5 | / | 16 | 0 | 0 |
|  | BERT-large | 1e-5 | 5 | / | 24 | 0 | 0 |
| ALBERT | ALBERT-base | 2e-5 | / | 12000 | 32 | 1000 | 0 |
|  | ALBERT-large | 1e-5 | / | 12000 | 32 | 1000 | 0 |
| RoBERTa | RoBERTa-base | 1e-5 | 4 | / | 16 | 1200 | 0.1 |
|  | RoBERTa-large | 1e-5 | 4 | / | 16 | 1200 | 0.1 |

Table A2: Hyperparameters for fine-tuning models that integrate with annotated rationales. We performed a 10-fold cross validation and early stopping based on the loss on development set. All modules were fine-tuned for fine-tuned models, while only the final linear feedforward layer was fine-tune for pretrained model.

|  | models | learning rate | training epochs | training batch size | weigh decay | early stopping patience |
|---|---|---|---|---|---|---|
| Fine-tuned | BERT, RoBERTa, and ALBERT | 1e-5 | 10 | 16 | 0.1 | 2 |
| Pretrained | BERT and RoBERTa | 1e-2 | 100 | 80 | 0 | 5 |
|  | ALBERT | 1e-3 | 100 | 80 | 0 | 5 |

## A.2 Additional plots

Figure A1: Performance of different models for Title, Purpose, and Cause questions.

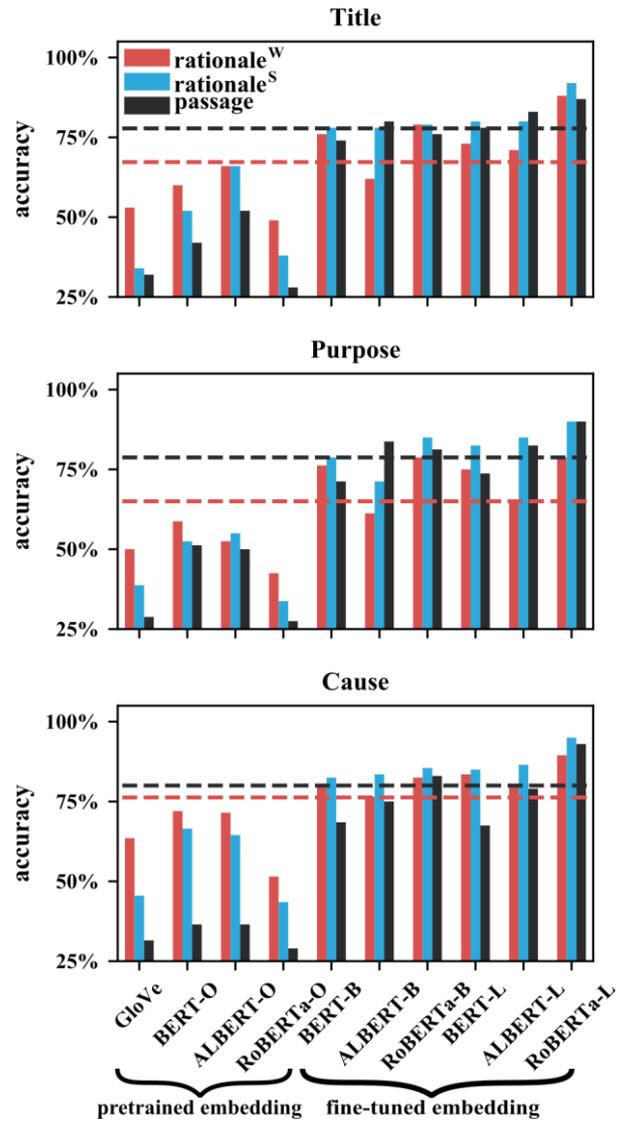